\documentclass[a4paper, 10pt, conference]{ieeeconf}

\newif\ifoagmfinalcopy

\oagmfinalcopytrue  

\IEEEoverridecommandlockouts                       

\usepackage{graphics} 
\usepackage{epsfig} 
\usepackage{mathptmx} 
\usepackage{times} 
\usepackage{amsmath} 
\usepackage{amssymb}  

\usepackage{lineno}
\usepackage{tikzpagenodes}
\usepackage{background}
\usepackage{gensymb}
\usepackage{subcaption}
\usepackage{tensor}
\usepackage{hyperref}
\usepackage{booktabs}
\usepackage[binary-units=true]{siunitx}

\usepackage[utf8]{inputenc}

\usepackage{pgfplots}
\usepackage{tikz-3dplot}

\usetikzlibrary{calc}
\usetikzlibrary{arrows}
\usetikzlibrary{angles}
\usetikzlibrary{quotes}
\usetikzlibrary{decorations.pathreplacing}
\usetikzlibrary{bending}
\usetikzlibrary{calc}

\newcommand{\icolvec}[1]{
    \left[\begin{smallmatrix}#1\end{smallmatrix}\right]^T
}

\ifoagmfinalcopy
\backgroundsetup{color=white}
\else

\linenumbers

\newcommand{\MyOAGMConfidentialLogo}{
\begin{tikzpicture}[remember picture,overlay]
\node[align=center,text=blue] at ([yshift=1em]current page text area.north) {\Large \#\#\# OAGM 2018 SUBMISSION: CONFIDENTIAL REVIEW COPY \#\#\#};
\end{tikzpicture}%
}

\SetBgContents{\MyOAGMConfidentialLogo}
\SetBgPosition{current page.north west}
\SetBgOpacity{0.5}
\SetBgAngle{0.0}
\SetBgScale{1.0}

\fi

\title{\LARGE \bf Large Area 3D Human Pose Detection Via Stereo Reconstruction in Panoramic Cameras}

\ifoagmfinalcopy
\author{Christoph Heindl$^{1}$, Thomas P\"{o}nitz$^{1}$, Andreas Pichler$^{1}$, and Josef Scharinger$^{2}$
\thanks{$^{1}$Profactor GmbH, Im Stadtgut A2, 4407 Steyr-Gleink, Austria}%
\thanks{$^{2}$JKU Department of Computational Perception, Altenbergerstr. 69, 4040 Linz, Austria}%
}
\else
\author{Anon, Ymous}
\fi

\begin{document}

\maketitle


\begin{abstract}
    We propose a novel 3D human pose detector using two panoramic cameras. We show that transforming fisheye perspectives to rectilinear views allows a direct application of two-dimensional deep-learning pose estimation methods, without the explicit need for a costly re-training step to compensate for fisheye image distortions. By utilizing panoramic cameras, our method is capable of accurately estimating human poses over a large field of view. This renders our method suitable for ergonomic analyses and other pose based assessments.
\end{abstract}

\section{INTRODUCTION AND RELATED WORK}
    \label{sec:intro} 
    Human pose estimation, characterized as the problem of localizing specific anatomic keypoints, has enjoyed substantial attention in recent years due to the large number of potential applications. It has been shown that keypoint based pose descriptions provide important cues for a variety of tasks such as activity recognition \cite{aggarwal2011human} and biomechanical analysis \cite{moeslund2006survey}. 
    
    Inferring pose from a highly articulated, potentially self-occluding, non-rigid body is, in general, a hard and ill-posed problem. Non-optical approaches encompass electromechanical \cite{motion2004gypsy} or inertial sensor \cite{roetenberg2009xsens} based suits. Optical methods traditionally applied intrusive active or passive markers \cite{vicon,qualisys} for keypoint detection. Early marker-free methods detected body parts in single images \cite{felzenszwalb2005pictorial, andriluka2009pictorial, pishchulin2013poselet, yang2013articulated}. 3D stereo imaging was used to infer human poses from sparse depth-maps \cite{yang2007reconstruction, matsumoto2000algorithm}. Real-time dense depth cameras greatly simplified the reconstruction task \cite{knoop2006sensor, zhu2007constrained,plagemann2010real,shotton2011real} by providing additional metric constraints.
    
    Recently, single and multi-person pose estimation in monocular images made significant progress \cite{pfister2015flowing, iqbal2016pose, iqbalMG16}. Especially the existence of large-scale human annotated datasets \cite{andriluka20142d, linMBHPRDZ14} accelerated deep learning based approaches \cite{simonyanZ14a,toshevS13,cao2017realtime}.
    
    Fisheye lenses have, despite their large field of view, received little attention mostly due to their inherent image distortions which significantly alter the appearance of objects as they move through its line of sight. Among the methods published, researchers have considered single person \cite{otsuka2008realtime} detection, safe human-robot interactions \cite{cervera2008safety} and head pose tracking \cite{stiefelhagen2000simultaneous}. As most of the optical solutions mentioned above assume a pinhole lens model, their results cannot be directly applied to fisheye images.

    In this work we propose a 3D pose detector using two fisheye cameras in general position. We apply a deep convolutional network based 2D pose estimator to the input images and reconstruct the corresponding 3D joint coordinates via stereoscopic constraints. We show that proper rectilinear view generation from raw fisheye input images allows us to avoid tedious dataset generation and network training steps. We demonstrate the usefulness of our approach in a challenging 6$\times$6 meter working area, and consider the applicability to ergonomic analysis with respect to accuracy and robustness. To our knowledge we are the first to propose a practical large-scale 3D human pose estimation system based on fisheye lenses.

    \begin{figure} [!t]
        \centering
        \includegraphics[width=0.98\columnwidth]{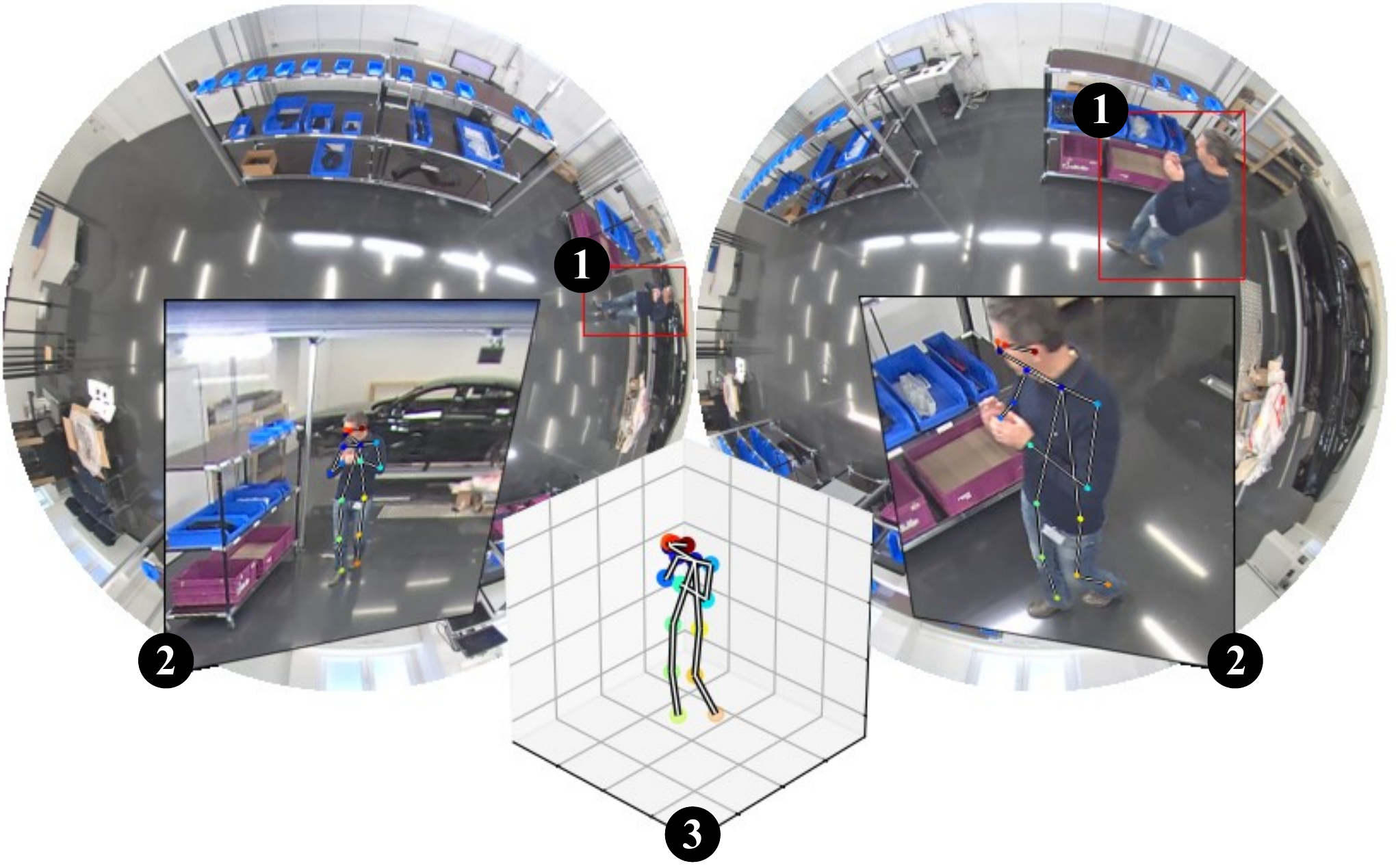}
        \caption {
            \label{fig:overview} 
            Overview. From two highly distorted \SI{180}{\degree} fisheye images coarse human location cues are inferred (1). Regions of interest are transformed into rectilinear views and articulated 2D human poses are then predicted via deeply learned architectures (2). The corresponding 2D joints are then triangulated via stereoscopic constraints to yield accurate 3D body part locations (3) even in the outer edges of a fisheye lens.
        }
    \end{figure}

\section{NOTATION}
    Throughout this work we use lower-case non-bold characters $x$ to denote scalars, bold-faced lower-case characters $\mathbf{x}$ represent column-vectors and upper-case bold characters $\mathbf{A}$ for matrices. $\mathbf{x}_i$ denotes the $i$-th element of $x$, $\mathbf{A}_{ij}$ the $i$-th row and $j$-th column of $\mathbf{A}$. For low dimensional vectors we also write $\mathbf{q}_x$ instead of $\mathbf{q}_0$ when it seems practical. 

\section{LENS MODELS AND PROJECTION FUNCTIONS}
    \label{sec:lensmodels} 
    A majority of pose estimation methods mentioned in Section \ref{sec:intro} assume that the camera model is sufficiently described by the pinhole camera model. Significant attention has therefore been paid in describing and correcting distortions that usually appear in ordinary lenses with moderate radial distortion \cite{zhang2000flexible, brown1966decentering}. However, these models are incompatible with wide angle lenses as their projective properties are not well captured. 
    
    We provide a brief overview of common lens models and projection functions in the next subsections. For an in depth discussion see \cite{basu1995alternative, schneider2009validation, hughes2010accuracy}. We consider only rotational symmetric lenses and assume that the principal point and the focal length is known. Both parameters can be determined by a number of methods \cite{kannala2006generic, shah1996intrinsic, gennery2006generalized}.

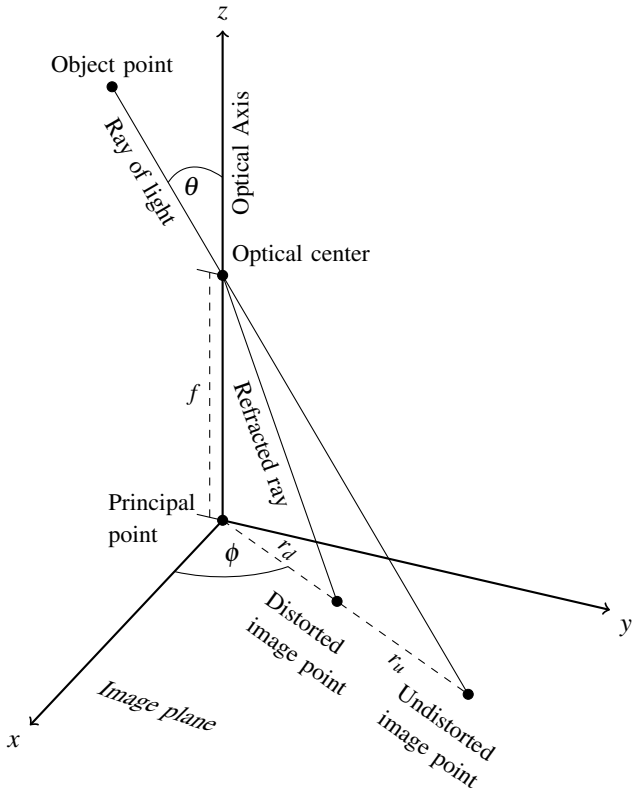
\begin{figure}[htp]
        \centering
        \tdplotsetmaincoords{60}{115}
        \begin{tikzpicture}[scale=7.5, tdplot_main_coords]

        \coordinate (o) at (0,0,0);
        \coordinate (z) at (0,0,1);
        \coordinate (oc) at (0,0,0.5);   
        \tdplotsetcoord{op}{.8}{25}{-120}
        \tdplotsetcoord{uip}{0.75116702}{90}{60}
        \tdplotsetcoord{dip}{0.35}{90}{60}

        \draw[thick,->] (0,0,0) -- (0.8,0,0) node[anchor=north east]{$x$};
        \draw[thick,->] (0,0,0) -- (0,0.75,0) node[anchor=north west]{$y$};
        \draw[thick,->] (0,0,0) -- (z) node[anchor=south]{$z$} ;

        \filldraw (op) circle[radius=0.25pt] node[anchor=south]{\small Object point};
        \filldraw (o) circle[radius=0.25pt] node[anchor=east, inner sep=10, align=left]{\small Principal \\ \small point};
        \filldraw (oc) circle[radius=0.25pt] node[anchor=south west]{\small Optical center};
        \filldraw (uip) circle[radius=0.25pt];
        \filldraw (dip) circle[radius=0.25pt]; 

        \path (o) -- (z) node[near end, below, sloped] {\small Optical Axis};
        \draw(op) -- (uip) node[very near start, below, sloped] {\small Ray of light};
        \draw (oc) -- (dip) node[midway, below, sloped] {\small Refracted ray};
        \draw[dashed] (o) -- (uipxy);
        \path (o) -- (dipxy) node[midway, above, sloped, inner sep=1.5] {\small $r_d$};
        \path (o) -- (uipxy) node[near end, below, sloped] {\small $r_u$};
        \path (o) -- (dipxy) node[at end, sloped, below, align=left, inner sep=10] {\small Distorted\\ \small image point};
        \path (o) -- (uipxy) node[at end, sloped, below, align=left, inner sep=10] {\small Undistorted\\ \small image point};
        \path (0.7, 0.1, 0.0) -- (0.8, 0.7, 0.0) node[very near start, sloped, above, align=left, xslant=0.5] {\small Image plane};

        \draw[thin] (oc) -- (0, -0.05, 0.5);
        \draw[thin] (o) -- (0, -0.05, 0.0);
        \draw[dashed] (0,-0.025,0) -- (0,-0.025, 0.5) node [midway, left] {\small $f$};

        \tdplotdrawarc{(o)}{0.2}{0}{60}{anchor=south}{$\phi$}
        \tdplotsetthetaplanecoords{-120}
        \tdplotdrawarc[tdplot_rotated_coords]{(oc)}{0.2}{0}{56.3510}{anchor=north}{$\theta$}

        \end{tikzpicture}

        \caption{\label{fig:lensmodel}
            Illustration of the general projection model and its related parameters. A ray of light emitted from a 3D object passes through the optical center and is potentially refracted due to lens characteristics. The measures $r_u$ and $r_d$ correspond to the ideal (rectilinear) and distorted (actual) radial distances from the principal point. The inclination angle $\theta$ measures the angular difference between the ray of light and the optical axis. The angle $\phi$ denotes the plane-polar angle around the optical axis. The focal length $f$ represents the distance between the optical center and the image plane.
        }
        
    \end{figure}

    We consider the characteristics of a lens to be captured in functional relationship between distorted image points on the focal plane and corresponding object points. As illustrated in Figure \ref{fig:lensmodel}, the radial distance $r_d$ of the optical center to the distorted image point is a function of the object's inclination angle with the z-axis $\theta$, the plane-polar angle around the optical axis $\phi$ and the focal length $f$. We consider radially symmetric lenses and therefore assume that the angle $\phi$, in contrast to $\theta$, remains unchanged by the lens (see Figure \ref{fig:lensmodel}).

    \subsection{Rectilinear projection}
        The most frequently found projection in computer vision is the pinhole projection. Due to the property that this projection preserves straight lines it is also termed the rectilinear projection. The projection function is given by 
        \begin{equation}
            \label{eq:rectilinear}
            r_d = f \tan(\theta).
        \end{equation}
        For this particular model $r_d = r_u$ and for large field of views the projected image becomes increasingly large and finally infinite when the field of view reaches \ang{180} degrees.

    \subsection{Fisheye projections}
        \label{sec:fisheyeprojections}
        Similar to rectilinear lenses, fisheye lenses have been manufactured to adhere to optical-engineered projection behavior. The projection functions governing these designs are also known as the classic projection functions and are listed below
        \begin{align}
            \text{Equidistant: } r_d &= f \theta \label{eq:equidistant}\\
            \text{Stereographic: } r_d &= 2f\tan(\tfrac{\theta}{2}) \label{eq:stereographic}\\
            \text{Equisolid: } r_d &= 2f\sin(\tfrac{\theta}{2}) \label{eq:equisolid}\\
            \text{Orthographic: } r_d &= f\sin(\theta) \label{eq:orthographic}.
        \end{align}

\begin{figure} [htp]
            \centering
            \includegraphics[width=0.98\columnwidth]{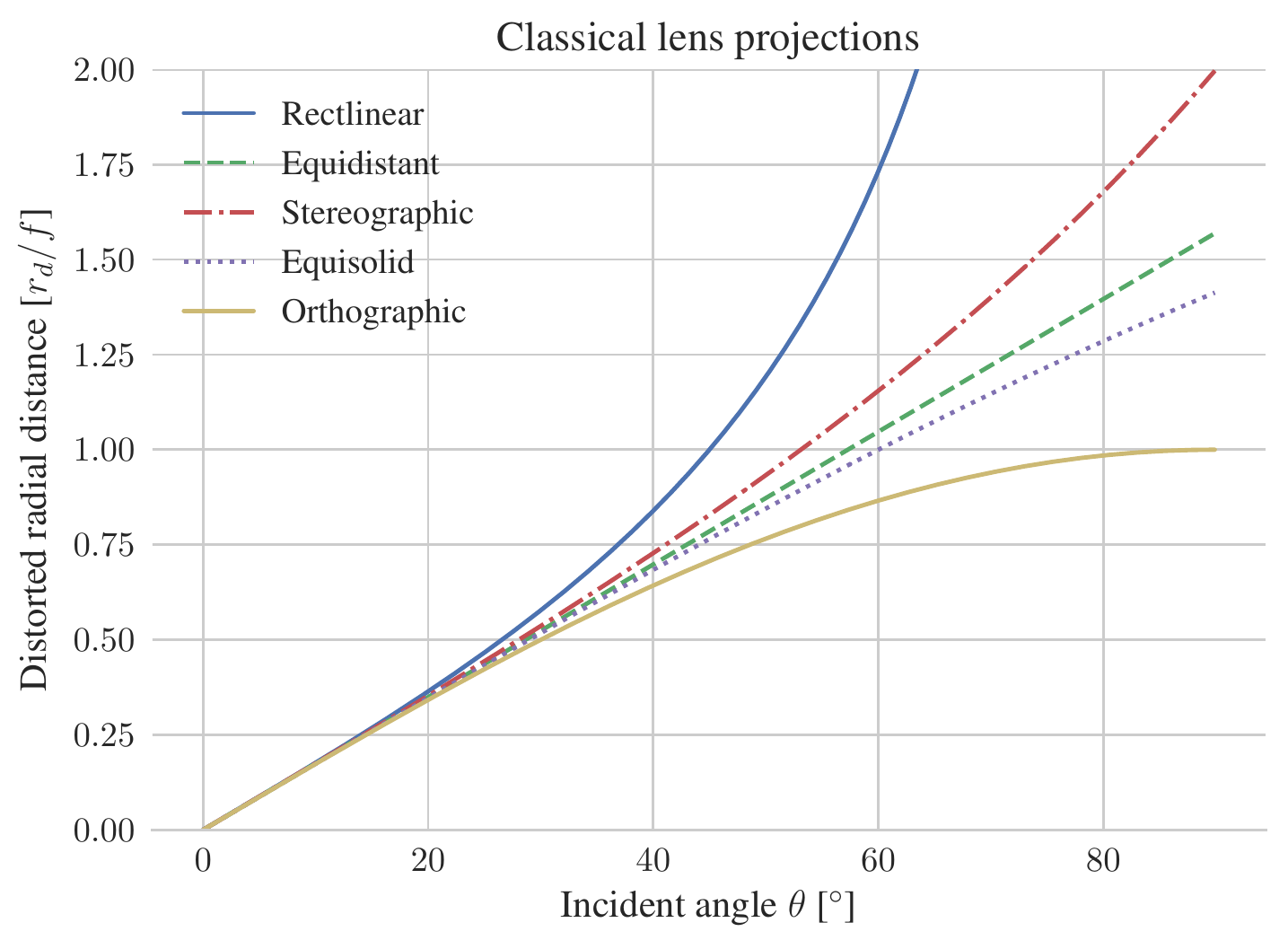}
            \caption {
                \label{fig:fisheyeprojections} 
                Plots of classical fisheye projection equations showing normalized radial distorted distances $r_d/f$ as a function of the inclination angle $\theta$. Note the monotonicity of the functions that ensures that all angles are mapped to different radial distances.
            }
        \end{figure}

        In Figure \ref{fig:fisheyeprojections} we compare radial distorted distances, $r_d$, as a function of the inclination angle, $\theta$, for all projection models. Note, the monotonicity of the functions that ensures that all angles are mapped to different radial distances.

        Besides the classic projection functions, various other models have been proposed. Most notably are polynomial models \cite{basu1995alternative}, a summation of sine terms model \cite{herbert1986calibration} and a universal model \cite{gennery2006generalized}. Unlike the classical optical-engineered models, these models try to capture a variety of different lenses in a single formula. While the classic projection formulas can be inverted algebraically (i.e determining the $\theta$ from $r_d$) this may not be as easily true for the alternatives.

\section{RECTILINEAR VIEW GENERATION}
\begin{figure} [htp]
        \centering
        \includegraphics[width=0.95\columnwidth]{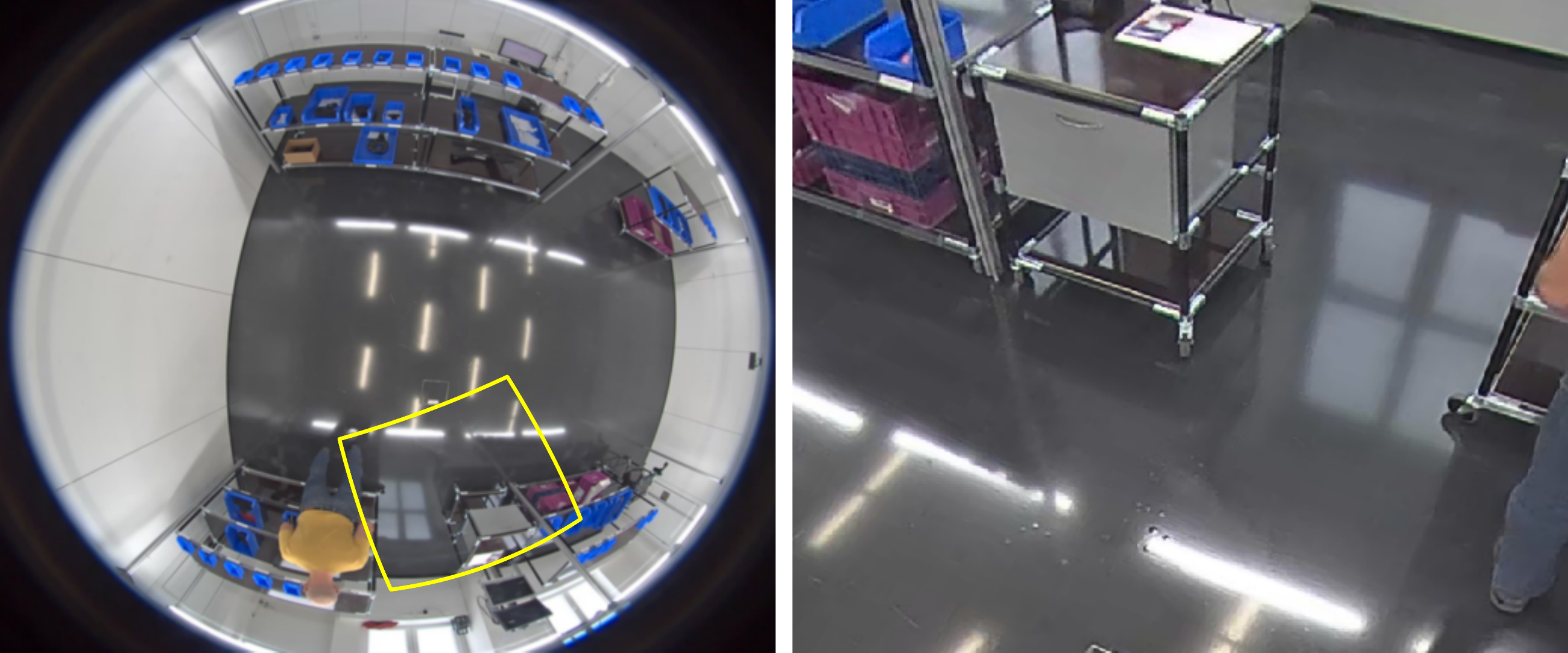}
        \caption {
            \label{fig:rectilinearview} 
            Upright rectilinear view generation. Fisheye input image on the left, rectilinear view on the right. The bounds of the rectilinear view are shown in yellow in the left image.
        }
    \end{figure}

    Generating rectilinear views from fisheye images, as shown in Figure \ref{fig:rectilinearview}, is an important technique in our approach, as our 2D pose detectors assume upright images taken by a pinhole camera model.

    Our approach to generate rectilinear views is based on virtual pinhole cameras that share the same origin as the fisheye cameras but are arbitrarily rotated with respect to their physical counterpart. In order to map image points between the artificial pinhole and physical fisheye view the following steps are applied in order.
    \begin{enumerate}
        \item \emph{Un-project} - computes object points from distorted image points using the destination camera model.
        \item \emph{Rotate} - rotates object points into the source camera space.
        \item \emph{Project} - computes distorted image points from object points using the source camera model.
    \end{enumerate}
    When applied to all pixels of a destination view, this method leads to efficient lookup maps of corresponding locations with sub-pixel accuracy. The destination image is then formed by interpolating pixels via lookup coordinates. While we are usually interested in mapping fisheye image and rotated pinhole image coordinates, our method works for any pair of camera models.

    \newcommand{\slfrac}[2]{\left.#1\middle/#2\right.}
    \newcommand{\norm}[1]{\left\lVert#1\right\rVert}
    
    \subsection{Projection from object space}
        \label{sec:projection}
        To compute distorted homogeneous image coordinates $\mathbf{i}=\icolvec{\mathbf{i}_x, \mathbf{i}_y, 1}$ for a Cartesian point $\mathbf{o}=\icolvec{\mathbf{o}_x, \mathbf{o}_y, \mathbf{o}_z}$ in object space, we first compute its spherical coordinates with respect to the camera intrinsic frame
        \begin{align}
            \begin{bmatrix}
                r \\ \theta \\ \phi
            \end{bmatrix}
            =
            \begin{bmatrix}
                \norm{\mathbf{o}} \\
                \arccos{\left(\slfrac{\mathbf{o}_z}{\norm{\mathbf{o}}}\right)} \\
                \arctan2(\mathbf{o}_y, \mathbf{o}_x)
            \end{bmatrix}.
        \end{align}
        Next, we compute $r_d$ from $\theta$ according to the lens projection model (see \ref{sec:fisheyeprojections}). The vector $\icolvec{r_d & \phi}$ then denotes the polar coordinates of the distorted image point. The Cartesian coordinates are given by
        \begin{align}
            \begin{bmatrix}
                \mathbf{i}_x \\ \mathbf{i}_y \\ 1
            \end{bmatrix}
            =
            \begin{bmatrix}
                r_d\cos{\phi}\\
                r_d\sin{\phi}\\
                0
            \end{bmatrix}
            +
            \begin{bmatrix}
                \mathbf{c}_x\\
                \mathbf{c}_y\\
                1
            \end{bmatrix}
        \end{align}
        where $\icolvec{\mathbf{c}_x & \mathbf{c}_y & 1}$ denotes the camera principal point. Henceforth, we denote the projection operation $\operatorname{project}(\mathbf{o};M) \colon \mathbb{R}^3 \rightarrow \mathbb{P}^2$ as a functional mapping that takes three-dimensional object points, $\mathbf{o}$, to homogeneous two-dimensional image points $\icolvec{\mathbf{i}_x, \mathbf{i}_y, 1}$ using the lens model $M$.

    \subsection{Reverse projection from image space}
        Reversing the process outlined in Section \ref{sec:projection} is ambiguous, as depth is lost during projection and all locations along a ray project to the same image coordinates. For our purposes it suffices to un-project image coordinates to points on the unit sphere, as our consideration mainly involves purely rotated cameras. First, we compute the polar coordinate representation for the image point $i$
        \begin{align}
            \mathbf{n} &= \mathbf{i} - \mathbf{c} \\
            \begin{bmatrix}
                r_d \\ \phi
            \end{bmatrix}            
            &=
            \begin{bmatrix}
                \norm{\mathbf{n}} \\ 
                \arctan2(\mathbf{n}_y, \mathbf{n}_x)
            \end{bmatrix}.
        \end{align}
        We then apply the reverse projection function to obtain $\theta$ from $r_d$ according to the lens model. The vector $\icolvec{r & \theta & \phi}$ describes a ray from the origin into object space through $i$ in spherical coordinates. Setting $r=1$, constrains the point to the unit sphere. Converting back to Cartesian coordinates gives
        \begin{align}
            \begin{bmatrix}
                \mathbf{o}_x \\
                \mathbf{o}_y \\
                \mathbf{o}_z 
            \end{bmatrix}
            =
            \begin{bmatrix}
                r\sin(\theta)\cos{\phi}\\
                r\sin(\theta)\sin{\phi}\\ 
                r\cos(\theta)
            \end{bmatrix}.
        \end{align}
        We define the reverse projection operation $\operatorname{unproject}(\mathbf{o};M) \colon \mathbb{P}^2 \rightarrow \mathbb{R}^3$ to be a functional mapping from  homogeneous image points $\icolvec{\mathbf{i}_x, \mathbf{i}_y, 1}$, to three-dimensional object points $\mathbf{o}$  using the lens model $M$.

        The general mapping of image points between two purely rotated cameras can now be written as 
        \begin{align}
            \mathbf{o} &= \operatorname{unproject}(\icolvec{\mathbf{i}_x & \mathbf{i}_y & 1};M) \\
            \mathbf{o'} &= \mathbf{R'}^T\mathbf{R}\mathbf{o} \\
            \mathbf{i'} &= \operatorname{project}(\mathbf{o'};M')
        \end{align}
        where $M$, $M'$ are the respective camera lens models and $\mathbf{R}$, $\mathbf{R'}$ are camera orientations. The mapping is simplified when both lens models are rectilinear, in which case the mapping can be conveniently described by a homography of the following form
        \begin{align}
            \mathbf{i'} = \mathbf{K'}\mathbf{R'}^T\mathbf{R}\mathbf{K}^{-1}\icolvec{\mathbf{i}_x, \mathbf{i}_y, 1}
        \end{align}
        where $\mathbf{K}$ and $\mathbf{K'}$ contain camera intrinsics.

\section{HUMAN POSE ESTIMATION}
    Human body part estimation consists of two steps. First, two-dimensional human poses are estimated in rectilinear views formed from both fisheye cameras. Then, a three-dimensional reconstruction is computed based on stereographic constraints. Both steps are detailed below.

    \subsection{2D Human Pose Estimation}
\begin{figure}[htp]
            \centering
            \begin{subfigure}{0.45\columnwidth}
                \centering
                \includegraphics[width=\columnwidth]{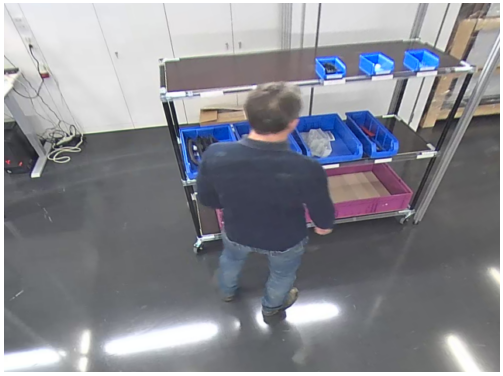}
                \caption{Rectilinear input image.}
            \end{subfigure}
            \hfill
            \begin{subfigure}{0.45\columnwidth}
                \centering
                \includegraphics[width=\columnwidth]{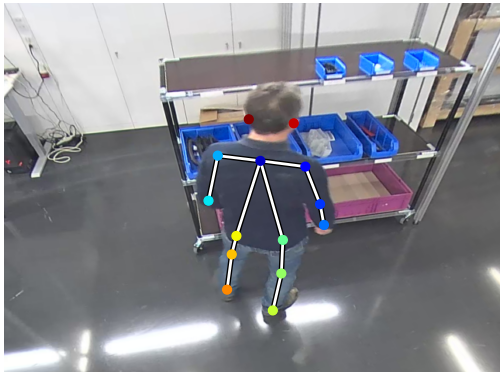}
                \caption{Computed body joints and limbs.}
            \end{subfigure}
            \\
            \begin{subfigure}{0.45\columnwidth}
                \centering
                \includegraphics[width=\columnwidth]{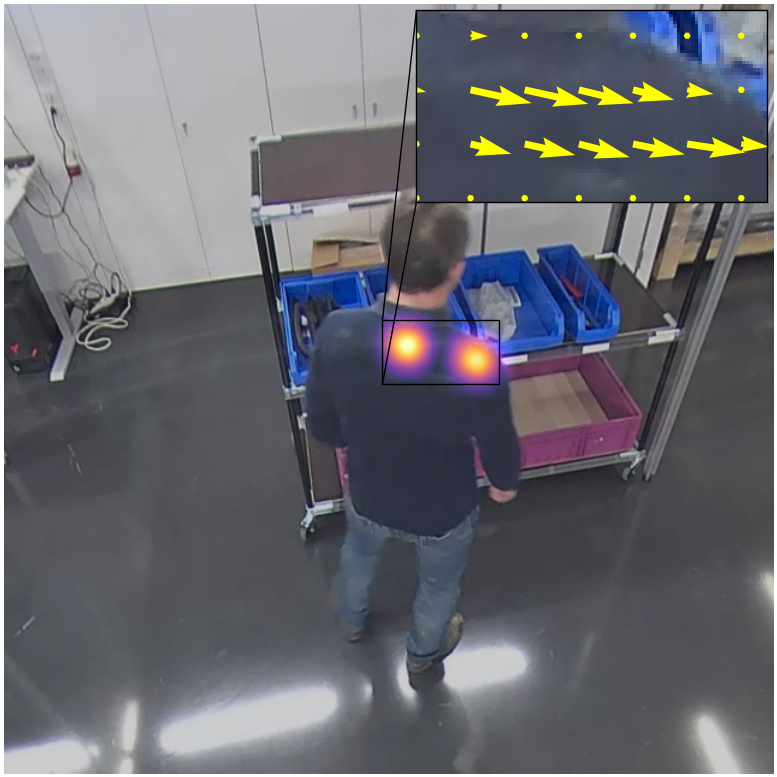}
                \caption{Joint confidences and part affinities between neck and right shoulder.}
            \end{subfigure}
            \hfill
            \begin{subfigure}{0.45\columnwidth}
                \centering
                \includegraphics[width=\columnwidth]{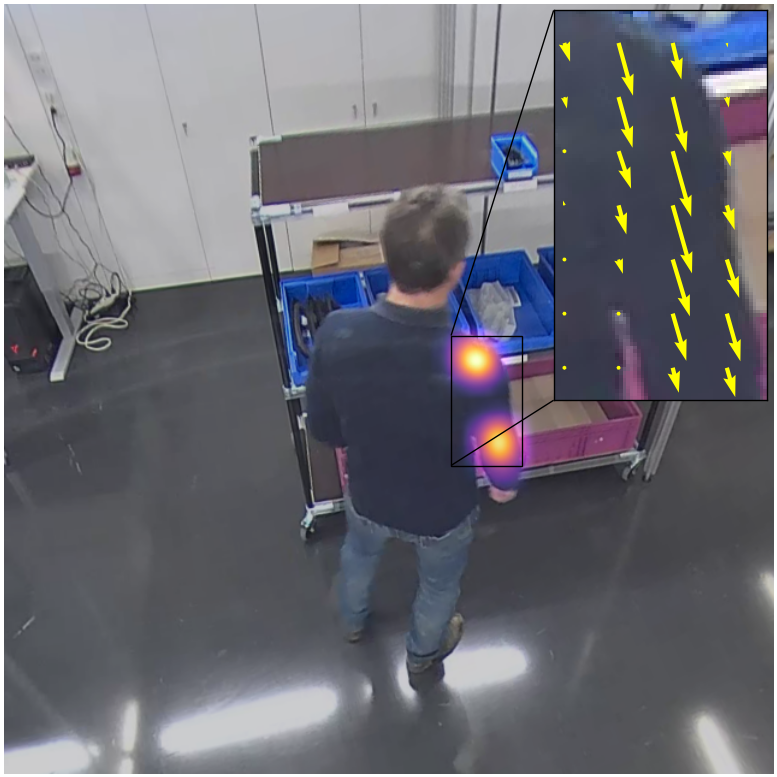}
                \caption{Joint confidences and part affinities between right shoulder and right elbow.}                
            \end{subfigure}
            \caption{
                \label{fig:2dposeestimation}
                Various stages of the 2D pose estimation algorithm. From a rectilinear view (top-left) a convolutional neural network predicts for every joint and limb a confidence and affinity vector field (bottom-right and-bottom left). A graph based algorithm constructs a skeleton body model based on these inputs (top-right).
            }
        \end{figure}

        Our 2D pose detection is based on the method of Cao et al. \cite{cao2017realtime}, which takes as input a rectilinear image view and outputs anatomic keypoint locations. A neural network is used to predict body joint confidence maps and a set of two dimensional vector fields that encode so called body limb affinities. The basic building block of the neural network is a set of convolutional filters that are iteratively applied to previous results in order to refine confidence and affinity maps. In the predicted confidence maps, peak localization is performed to identify potential joint candidates. Then, a graph algorithm guided by affinity maps is used to determine the connectivity. Figure \ref{fig:2dposeestimation} illustrates various stages of the algorithm.
        
        As the detector is directly applied to upright rectilinear views we can use pre-trained network weights and avoid time consuming manual annotation of fisheye images. In order to bootstrap the rectilinear view generation an initial guess of people positions in fisheye images is needed. This can be solved in a number of ways, such as using a people detector \cite{wojke2017simple} or performing foreground segmentation \cite{kim2005real}. Once an initial view orientation is known, subsequent rectilinear views can be computed automatically by re-focusing the view on detected 2D human poses.

    \subsection{3D Human Pose Reconstruction}
        Computing 3D joint coordinates requires at least 2 fisheye images with projections of the same world space joint. Given a rigid transformation between two capture devices, the 3D location can be obtained via triangulation. For static camera setups the rigid transformation can be estimated \cite{abraham2005fish} in a preprocessing step. For non-rigid setups, camera position and scene geometry needs to be inferred simultaneously. This is considered a bundle adjustment problem for which numerous iterative non-linear solutions have been proposed \cite{hartley2003multiple, ma2012invitation}. 
        
\begin{figure*}
            \centering
            \begin{subfigure}[b]{0.65\textwidth}
                \centering
                \includegraphics[width=\textwidth]{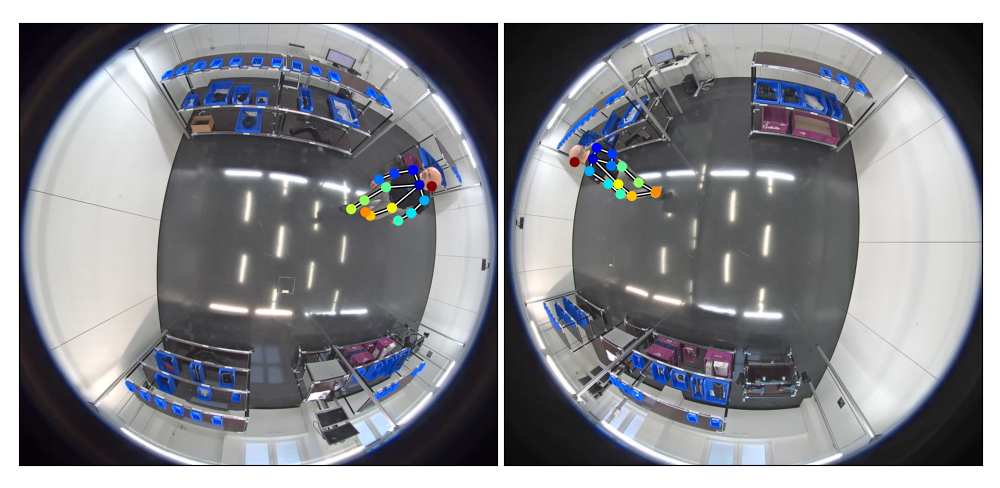}
            \end{subfigure}
            \begin{subfigure}[b]{0.33\textwidth}
                \centering
                \includegraphics[width=\textwidth]{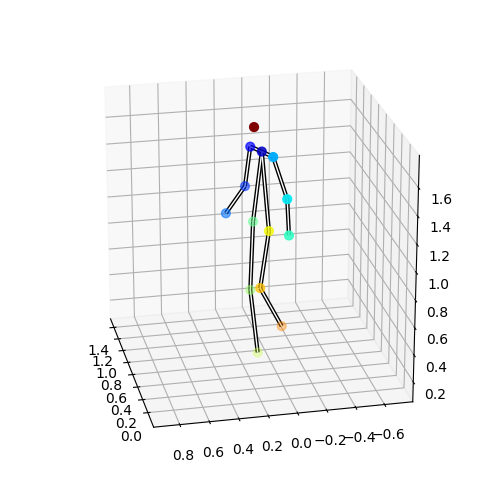}
            \end{subfigure}
            \\
            \begin{subfigure}[b]{0.65\textwidth}
                \centering
                \includegraphics[width=\textwidth]{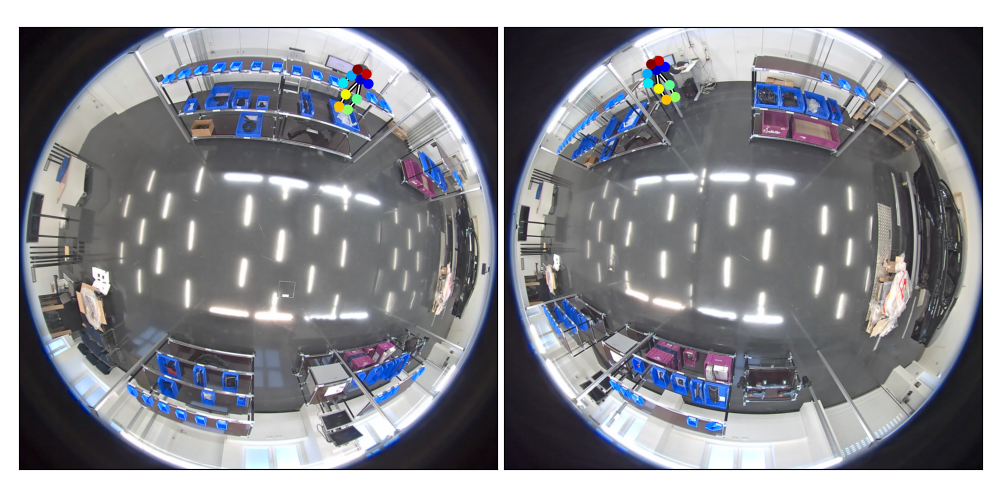}
            \end{subfigure}
            \begin{subfigure}[b]{0.33\textwidth}
                \centering
                \includegraphics[width=\textwidth]{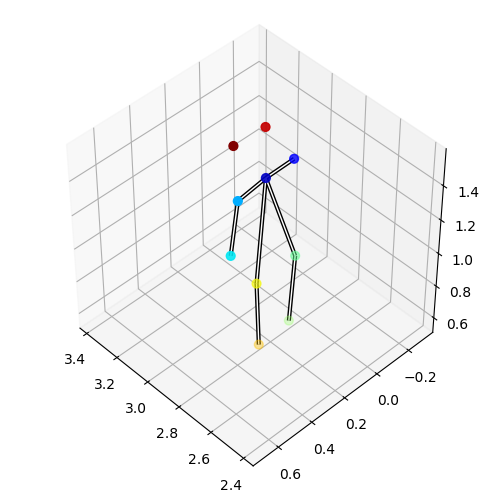}
            \end{subfigure}
            \\          
            \begin{subfigure}[b]{0.65\textwidth}
                \centering
                \includegraphics[width=\textwidth]{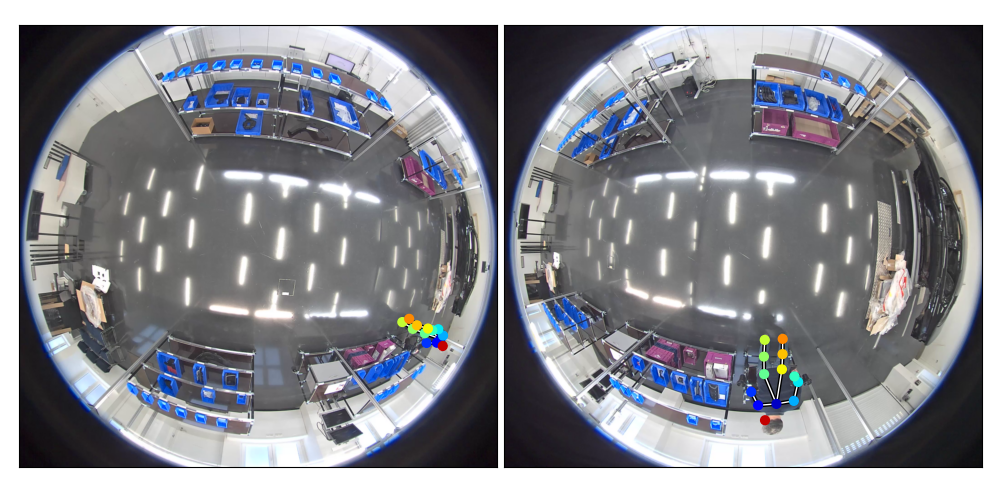}
            \end{subfigure}
            \begin{subfigure}[b]{0.33\textwidth}
                \centering
                \includegraphics[width=\textwidth]{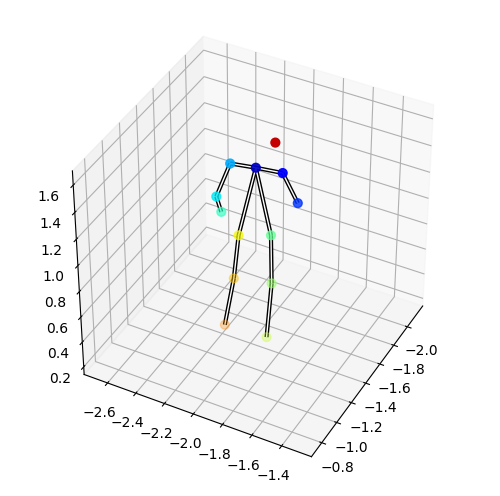}
            \end{subfigure}
            \\     
            \caption{
                \label{fig:3dposeestimation}
                Examples of 3D stereo reconstruction of human body joints in fisheye images. Left/middle: input images and superimposed detected two dimensional body joints. Right: Three dimensional metric body model.
            }
        \end{figure*}

        In either case, the usual linear epipolar constraints for stereo setups do not hold, because fisheye cameras exhibit non-linear projection functions. Therefore, we perform triangulation directly in rectilinear views instead of fisheye images, for which the constraints hold. Without loss of generality, let $\mathbf{P} \in \mathbb{R}^{3 \times 4}$ be the camera projection matrix of the first rectilinear camera given by
        \begin{align}
            \label{eq:projectionmatrix}
            \mathbf{P} = \mathbf{K}\begin{bmatrix}\mathbf{I} & \mathbf{0}\end{bmatrix}\begin{bmatrix}\mathbf{R}^T & \mathbf{0}\\0 & 1\end{bmatrix}\mathbf{W}^{-1}
        \end{align}
        where $\mathbf{K} \in \mathbb{R}^{3 \times 3}$ is the rectilinear projection matrix, $\mathbf{R} \in \mathbb{R}^{3 \times 3}$ the orientation of rectilinear view with respect to the parental fisheye camera, $\mathbf{W} \in \mathbb{R}^{4 \times 4}$ is the position and orientation of the fisheye camera in world space and $\mathbf{I} \in \mathbb{R}^{3 \times 3}$ is the identity matrix. Similarily we define $\mathbf{P'}$ for the second rectilinear view. The simultaneous projection of a homogeneous world point $\mathbf{x}$ in either camera focal plane is given by 
        \begin{align}
            w\icolvec{\mathbf{i}_x, \mathbf{i}_y, 1} &= \mathbf{P}\mathbf{x} \label{eq:hprojection}\\
            w'\icolvec{\mathbf{i'}_x, \mathbf{i'}_y, 1} &= \mathbf{P'}\mathbf{x}.
        \end{align}
        Rewriting Equation \ref{eq:hprojection} line by line and denoting by $\mathbf{p}_i$ the i-th row of $\mathbf{P}$ we get
        \begin{align}            
            w\mathbf{i}_x &= \mathbf{p}_0\mathbf{x} \label{eq:hprojection1} \\
            w\mathbf{i}_y &= \mathbf{p}_1\mathbf{x} \label{eq:hprojection2}\\
            w &= \mathbf{p}_2\mathbf{x} \label{eq:hprojection3}.
        \end{align} 
        Then, the Direct Linear Transform (DLT)\cite{hartley1992stereo} algorithm is given by eliminating $w$ from Equations \ref{eq:hprojection1}, \ref{eq:hprojection2} via substitution using Equation \ref{eq:hprojection3}. Rewriting leads to two linear equations in four unknowns of $\mathbf{x}=\icolvec{x & y & z & w}$
        \begin{align}
            (x'\mathbf{p}_2-\mathbf{p}_0) \mathbf{x} &= 0 \\
            (y'\mathbf{p}_2-\mathbf{p}_1) \mathbf{x} &= 0.
        \end{align}
        Two views then yield the four equations. The system of equations may be written in matrix form as $\mathbf{A}\mathbf{x}=\mathbf{0}$ and solved for $\mathbf{x}$ in multiple ways \cite{hartley1997triangulation}. The DLT algorithm allows us to compute 3D point coordinates for corresponding image coordinates in two rectilinear views. Applying it to two-dimensional joint correspondences leads to three-dimensional reconstruction of body parts. Figure \ref{fig:3dposeestimation} shows several successful reconstructions.

\section{EVALUATION AND RESULTS}
    The setup we use for evaluation covers a 6$\times$6 meter working area with two fisheye cameras of type Axis M3007-PV mounted to the ceiling at height of 3 meters. The baseline between the cameras is roughly 1.5 meters. The simulated working area consists of several shelves that often cause partial body occlusions. We captured raw RGB video from both cameras at rate of 12 FPS at a resolution of 2592$\times$1944 over a period 4 weeks producing a total of 20 hours material. The video data contains 4 different people performing common assembly tasks. 
    
    The fisheye cameras have been intrinsically calibrated using the method described in \cite{kannala2006generic}. We obtained the extrinsic calibration for each camera separately by using an external tracking device\footnote{HTC Vive \url{ https://www.vive.com/eu/}}, whose tracking targets can be automatically detected in images. By capturing a set of 3D and corresponding distorted 2D image correspondences in rectilinear views, we solve for the unknown pose $\mathbf{W}$ using a iterative scheme \cite{levenberg1944method}. 
    
    We assess the accuracy of the calibration and rectilinear view generation by measuring lengths of known objects in reconstructed stereo scenes. For better readability we split the field of view of the fisheye camera into disjoint rings corresponding to increasing radial distortions. The results are shown in Table \ref{tab:accuracy}.
    \begin{table}[htp]
        \centering        
        \begin{tabular}{@{}lll@{}}
        \toprule
                                & Target length (\si{\meter}) & Measured length (\si{\meter})\\ 
        \midrule
        Central area   & 1.5                        & 1.49 $\pm$0.01             \\
        Outer area     & 1.5                        & 1.52 $\pm$0.018           \\
        Lens edge area & 1.5                        & 1.56 $\pm$0.035           \\ 
        \bottomrule
        \end{tabular}
        \caption{Measurement errors incurred by the stereo setup inaccuracies. For better comprehensibility we split the fisheye field of view into three concentric rings that mark central (low-error), outer area (mid-error) and edge area (high-error). Shown are target lengths as well as upper/lower limits over multiple measurements.} 
        \label{tab:accuracy}
    \end{table}

\begin{figure}[htp]
        \centering
        \includegraphics[width=0.7\columnwidth]{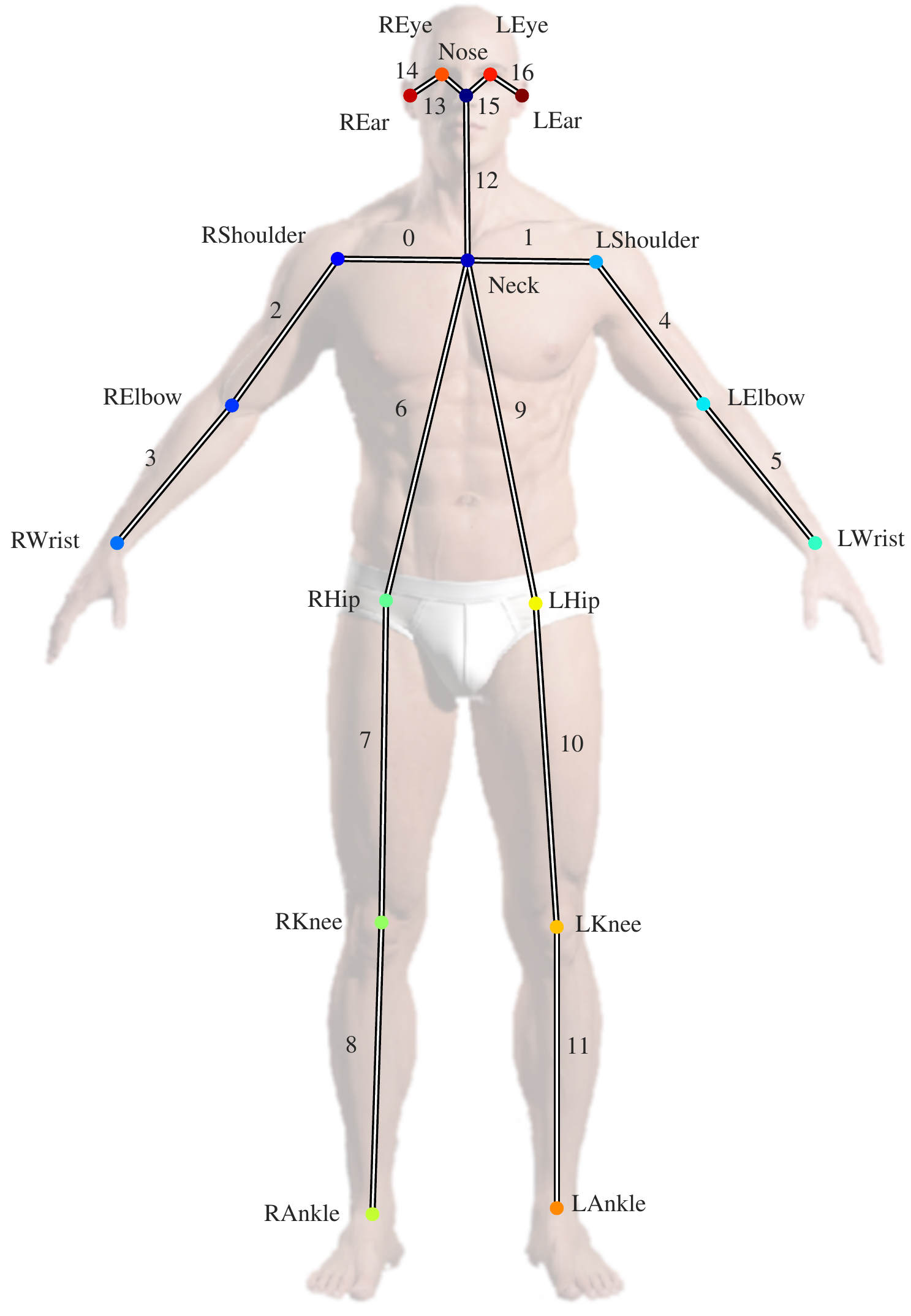}
        \caption{Joint names and limb indices for the COCO\cite{linMBHPRDZ14} model.}
        \label{fig:cocoann}
    \end{figure}

\begin{figure}[htp]
        \centering
        \begin{subfigure}[b]{0.98\columnwidth}
            \centering
            \includegraphics[width=\columnwidth]{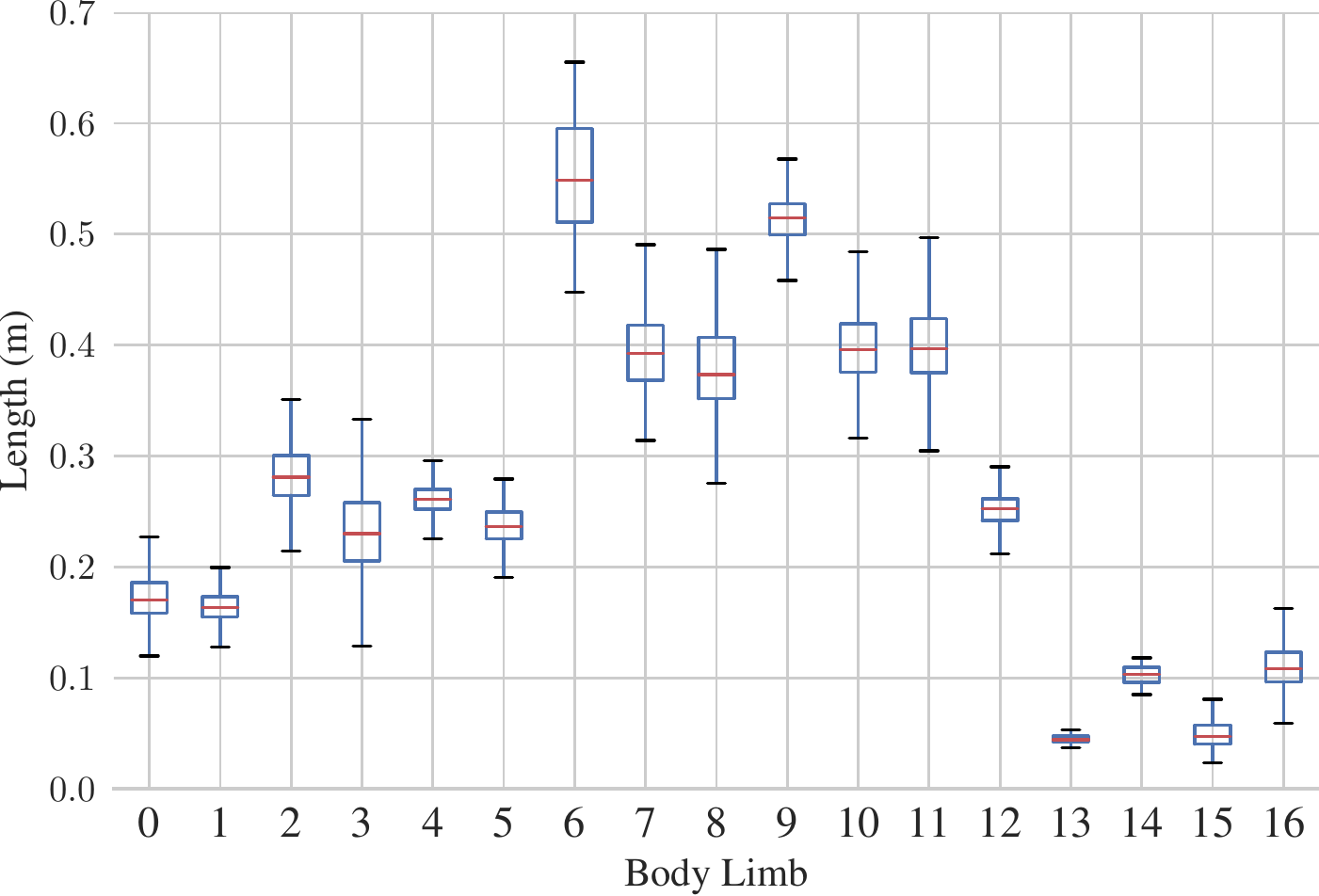}
        \end{subfigure}
        \bigskip
        \begin{subfigure}[b]{0.98\columnwidth}
            \centering
            \includegraphics[width=\columnwidth]{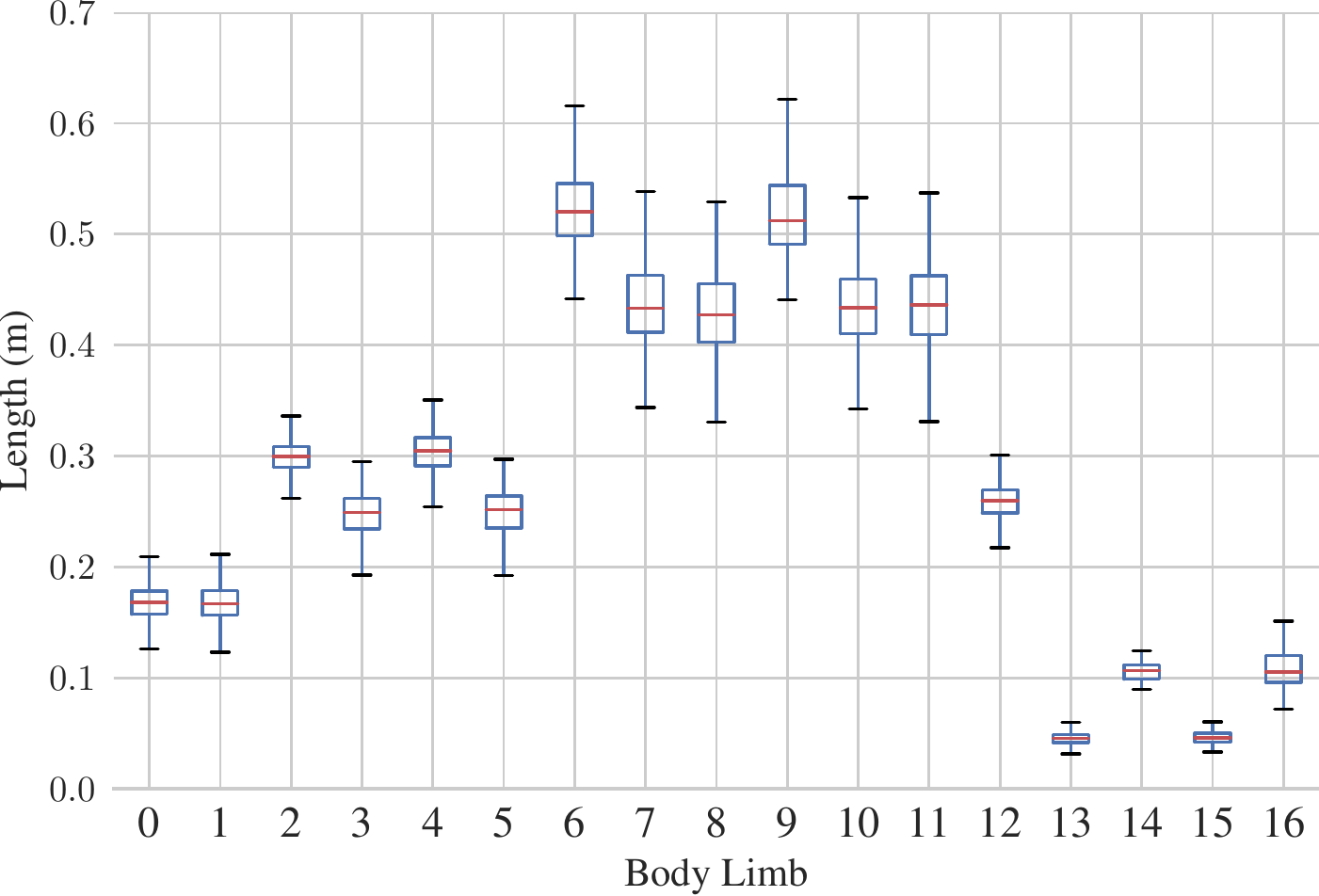}
        \end{subfigure}
        \caption{
            \label{fig:3dposeeval}
            Metric limb length statistics for two different people over sequence of 10800 frames (15 minutes). Note, the second person has longer legs - he is roughly 5-8 cm taller. The area covered is 6$\times$6 meter and includes serval obstacles that lead to occlusions. Refer to Figure \ref{fig:cocoann} for limb number lookup. 
        }
    \end{figure}

\begin{figure}[htp]
        \centering
        \begin{subfigure}[b]{0.98\columnwidth}
            \centering
            \includegraphics[width=\columnwidth]{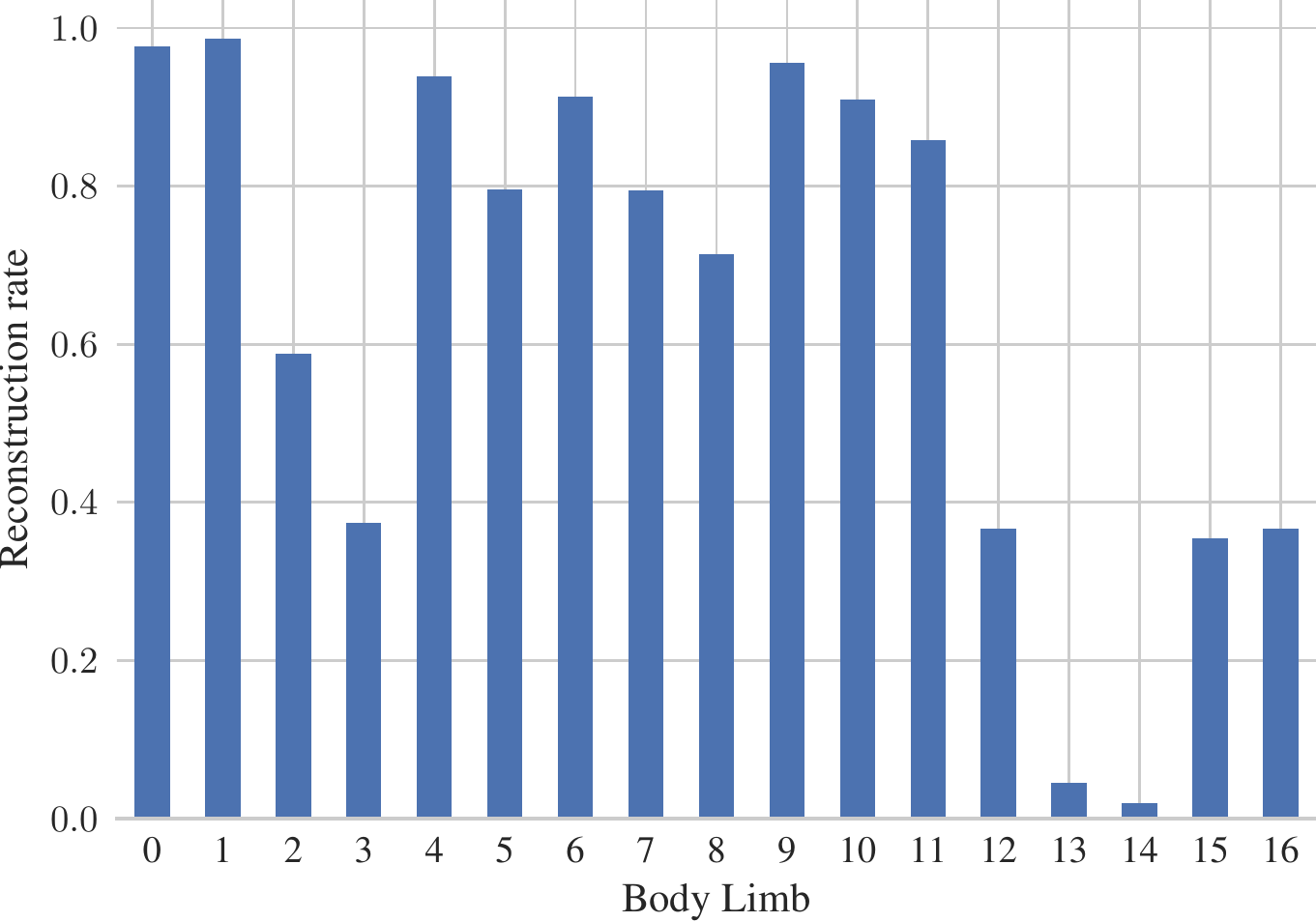}
        \end{subfigure}
        \caption{
            \label{fig:3dposeevalfreq}
            Reconstruction frequencies of individual limbs over the entire video testing set. Refer to Figure \ref{fig:cocoann} for limb number lookup. Facial features are reconstructed less frequently compared to larger body parts due to visibility constraints. 
        }
    \end{figure}

    We trained the 2D pose estimation algorithm on the COCO\cite{linMBHPRDZ14} dataset, which defines 18 body joints and 17 limbs (see Figure \ref{fig:cocoann}). Since the accuracy of the 2D pose estimation has already been studied elsewhere\cite{cao2017realtime, ning2017dual}, we concentrate on evaluating the quality of the 3D reconstruction. We validate the 3D reconstruction by accumulating 3D limb lengths over video segments grouped by individual persons. Ideally, limb lengths are stationary. However, due to stereo setup imprecision and fluctuations in 2D detection we observe varying limb lengths as shown in Figure \ref{fig:3dposeeval}. Bear in mind that the errors are mostly introduced when people move around in the lens edge area. Figure \ref{fig:3dposeevalfreq} shows the relative reconstruction frequencies of each limb.
    
    We ran the evaluation on a workstation with an Intel i7-7700 \SI{3.6}{\giga\hertz}, \SI{16}{\giga\byte} RAM, and a NVIDIA GeForce GTX1060 graphics card with \SI{6}{\giga\byte} memory. Relevant performance metrics for key stages in our algorithm are given in Table \ref{tab:timings}.    
    \begin{table}[htp]
        \centering        
        \begin{tabular}{@{}llll@{}}
        \toprule
        View                & View & 2D & 3D \\
        Resolution          & Generation (\si{\second}) & Detection (\si{\second}) & Reconstruction (\si{\second}) \\
        \midrule
        320$\times$320      & 0.02 $\pm$0.001 & 0.60 $\pm$0.02 & 0.01 $\pm$0.001  \\
        640$\times$640      & 0.10 $\pm$0.01  & 1.80 $\pm$0.05 & 0.01 $\pm$0.001  \\
        \bottomrule
        \end{tabular}
        \caption{Performance timings of key stages in our algorithm. We compare two different resolutions of rectilinear views and note how they affect each stage of the reconstruction pipeline.}
        \label{tab:timings}
    \end{table} 

\section{CONCLUSION AND FUTURE WORK}
In this paper, we proposed a novel human pose detector that predicts three-dimensional body part locations from two highly distorted fisheye cameras in general position. We demonstrated that the highly enlarged field of view of a fisheye lens is a compelling advantage in reducing hardware complexity. Especially the number of cameras needed to capture the scene can be reduced and thus many related calibration efforts can be avoided. With regard to pose evaluations, we find that analyses are possible with an accuracy of 2-3cm over a range of 6x6 meters.

We utilized recent deep-learning based approaches to 2D pose estimation in images and showed that generating artificial rectilinear views avoids the re-training of the neural network. To our knowledge we are the first to consider deep-learning based human pose reconstruction using stereo fisheye lenses. As a matter of fact we observe increasing inaccuracies in 3D reconstruction in the limit of the lens. 

In future work we will therefore reconsider the current triangulation method and verify whether additional smoothness constraints can help to reduce the errors. Another point of interest is reduction of runtime complexity, by improving the runtime of the 2D pose detector, in order to achieve real-time performance. 

\section*{ACKNOWLEDGMENT}
\ifoagmfinalcopy
    This research was supported in part by Lern4MRK (Austrian Ministry for Transport, Innovation and Technology), ”FTI Struktur
    Land Oberoesterreich (2017-2020)”, the European Union in cooperation with the State of Upper Austria within the project Investition in Wachstum und Beschäftigung (IWB), as well as AutoScan (FFG, 853416).
\else
    Acknowledgements have been removed for blind review purposes.
\fi

{
\small
\bibliographystyle{ieeetr}
\bibliography{3dstereoskeleton}
}

\end{document}